\documentclass[letterpaper]{article} 
\usepackage{aaai24}  
\usepackage{times}  
\usepackage{helvet}  
\usepackage{courier}  
\usepackage[hyphens]{url}  
\usepackage{graphicx} 
\usepackage{natbib}  
\usepackage{caption} 
\usepackage{algorithm}
\usepackage{algorithmic}

\usepackage{amsmath}
\usepackage{amssymb}
\usepackage{multirow}
\usepackage{graphicx}
\usepackage{subcaption}
\usepackage{color}
\usepackage{array}
\usepackage{url}
\usepackage{makecell}

\DeclareMathOperator*{\argmax}{arg\,max}

\usepackage{newfloat}
\usepackage{listings}
\DeclareCaptionStyle{ruled}{labelfont=normalfont,labelsep=colon,strut=off} 
\floatstyle{ruled}
\newfloat{listing}{tb}{lst}{}
\floatname{listing}{Listing}
\title{Improving Autonomous Separation Assurance through Distributed Reinforcement Learning with Attention Networks}

\author {
    Marc W. Brittain,
    Luis E. Alvarez,
    Kara Breeden
}
\affiliations {
    Massachusetts Institute of Technology Lincoln Laboratory\thanks{Distribution Statement A. Approved for public release. Distribution is unlimited. This material is based upon work supported by the United States Air Force under Air Force Contract No. FA8702-15-D-0001. Any opinions, findings, conclusions or recommendations expressed in this material are those of the author(s) and do not necessarily reflect the views of the United States Air Force. Delivered to the U.S. Government with Unlimited Rights, as defined in DFARS Part 252.227-7013 or 7014 (Feb 2014). Notwithstanding any copyright notice, U.S. Government rights in this work are defined by DFARS 252.227-7013 or DFARS 252.227-7014 as detailed above. Use of this work other than as specifically authorized by the U.S. Government may violate any copyrights that exist in this work.\newline \copyright~2023 Massachusetts Institute of Technology.}\\
    marc.brittain@ll.mit.edu, luis.alvarez@ll.mit.edu, kara.breeden@ll.mit.edu
}

\begin{document}

\maketitle

\begin{abstract}

Advanced Air Mobility (AAM) introduces a new, efficient mode of transportation with the use of vehicle autonomy and electrified aircraft to provide increasingly autonomous transportation between previously underserved markets. Safe and efficient navigation of low altitude aircraft through highly dense environments requires the integration of a multitude of complex observations, such as surveillance, knowledge of vehicle dynamics, and weather. The processing and reasoning on these observations pose challenges due to the various sources of uncertainty in the information while ensuring cooperation with a variable number of aircraft in the airspace. These challenges coupled with the requirement to make safety-critical decisions in real-time rule out the use of conventional separation assurance techniques. We present a decentralized reinforcement learning framework to provide autonomous self-separation capabilities within AAM corridors with the use of speed and vertical maneuvers. The problem is formulated as a Markov Decision Process and solved by developing a novel extension to the sample-efficient, off-policy soft actor-critic (SAC) algorithm. We introduce the use of attention networks for variable-length observation processing and a distributed computing architecture to achieve high training sample throughput as compared to existing approaches. A comprehensive numerical study shows that the proposed framework can ensure safe and efficient separation of aircraft in high density, dynamic environments with various sources of uncertainty.

\end{abstract}

\section{Introduction}
Advanced air mobility (AAM) is set to revolutionize transportation by introducing highly automated aircraft to transport passengers and cargo within local, regional, inter-regional, and urban environments~\cite{faa_conops}. However, the realization of AAM faces several key challenges, including safety, security, social acceptance, resilience, environmental impacts, regulation, scalability, and flexibility~\cite{NAP25646}.

To address these challenges, the use of advanced automation techniques such as artificial intelligence (AI) is essential. Specifically, learning-based decentralized separation assurance holds significant potential for enabling the safe and efficient operation of highly automated aircraft in the high-density, high-tempo AAM environment envisioned by the Federal Aviation Administration (FAA) and the National Aeronautics and Space Administration (NASA)~\cite{faa_conops,nasa_conops}.

However, due to the lack of real operational data and scenarios for AAM, developing, training, and validating AI algorithms becomes a challenge. Simulation provides a low-cost way to overcome this challenge by allowing for the exploration of edge-case scenarios that may be too dangerous to perform in the real world. Therefore, simulation-based training and validation of AI algorithms are essential for safe and efficient operation of AAM. Recently, the AI testbed for Advanced Air Mobility (AAM-Gym) was developed to provide a standardized ecosystem for the research of AI in AAM. By leveraging simulation backends such as BlueSky~\cite{bluesky} and UAMToolkit~\cite{doi:10.2514/6.2021-2381}, representative real-world scenarios can be developed for training and evaluation.

Recently, deep reinforcement learning (DRL) has demonstrated superior performance to humans in games such as Atari, GO, Warcraft, and StarCraft II, requiring a sophisticated balance between near-term and long-term strategic decisions~\cite{mnih2013playing, silver2016mastering, amato2010high, vinyals2017starcraft}. In addition, DRL has also been applied to air traffic control (ATC) and conflict resolution, where early work used an AI agent to mitigate conflicts and minimize the delay of aircraft reaching their metering fixes~\cite{HDRL}. In later works, \cite{pham2019machine} demonstrated that an AI agent can effectively resolve randomly generated conflict scenarios between a pair of aircraft through vectoring maneuvers. To encourage human ATC adoption of AI maneuvers, \cite{tran2020interactive} developed an interactive conflict solver using DRL that was trained using human resolution maneuvers, providing AI behavior more closely aligned with humans. More recently, \cite{ribeiro2020improvement, ribeiro2022improving} proposed a hybrid geometric-reinforcement learning algorithm for resolving conflicts in low-altitude airspace. \cite{badea2022lateral} explored the use of both lateral and vertical maneuvers for conflict resolution using DRL in traditional airspace. While these approaches are effective for sparse airspace environments, they fail to handle state space scalability as the number of intruder aircraft increases due to the either centralized, single-agent architectures or fixed-length state vectors with a maximum number of intruder aircraft.
In \cite{D2MA, D2MAV,D2MAV_A, MAASA}, it is shown how a decentralized separation assurance framework can alleviate the aforementioned scalability concerns and prevent loss of separation in high-density stochastic sectors by leveraging long short-term memory networks (LSTM) and attention networks, even when agents may be optimizing non-homogeneous reward functions.

In this article, a decentralized learning-based framework for aircraft separation assurance is introduced and applied to a high-density AAM use-case. We integrate the sample efficient Discrete Soft Actor-Critic (SACD) algorithm and extend the algorithm with the use of attention networks to handle a variable-length state space. Given SACD is an off-policy algorithm, an asynchronous training architecture is developed to decouple the agent-environment interaction with the algorithm training. This allows us to achieve an approximately 10x increase in the number of transitions trained over existing approaches. We show that the increased training leads to improved safety and operational suitability performance, even in highly uncertain environments.
The main contributions of this article are summarized as follows:
\begin{itemize}
\item We propose a scalable, distributed, and sample efficient aircraft separation assurance framework based on SACD and attention networks that is capable of both improving safety and operational suitability.
\item We introduce an expanded action set over prior works with the introduction of vertical maneuvers.
\item A representative AAM environment is developed in AAM-Gym, providing a comprehensive environment for evaluating the effectiveness of the proposed framework.
\end{itemize}

The structure of this paper is as  follows. We first provide a brief overview of reinforcement learning and soft actor-critic. Then, we introduce the approach to applying reinforcement learning to aircraft separation assurance. Following, details on the environment setup and numerical experiments are presented. We then discuss the results and summarize our findings in the conclusion.

\section{Background}
\label{background}
In this section, we briefly review the background of reinforcement learning and soft actor-critic.

\subsection{Reinforcement Learning}
Reinforcement learning (RL) is one type of sequential decision making where the objective is to learn a policy in a given environment. RL requires the environment to be formulated as a Markov Decision Process (MDP); a mathematical framework for modeling decision making processes with stochastic transitions. An MDP is defined by the tuple $(S,A,R,T,\gamma)$, where an agent in state $s \in S$ takes an action $a \in A$, transitions to state $s'$ with probability $T(s'|s,a)$, and receives a reward $R(s,a)$. In RL, the transition matrix $T$ is often unknown. The discount factor $\gamma$ determines how far in the future to look for rewards, where immediate rewards are emphasized as $\gamma\rightarrow$~0 and future rewards are prioritized when $\gamma\rightarrow$~1.

The RL agent is able to derive an optimal policy $\pi^{*}$ in the environment by maximizing a cumulative reward function
\begin{equation}
    \pi^{*} = \argmax_{\pi}E[\sum_{t=0}^{\tau}(r(s_{t}, a_{t})|\pi)],
\end{equation}
where $\tau$ represents the total time for a given environment. In environments with discrete or low-dimensional state-action representations, the optimal policy $\pi^{*}$ can be obtained using dynamic programming approaches such as Q-learning~\cite{watkins1992q}. However, many real-world environments can not be represented by discrete values or require high-dimensional state representations, requiring the use of function approximation for the policy. The aforementioned issues can be addressed through deep reinforcement learning (DRL) where a neural network is used to represent the policy $\pi$.

\subsection{Soft Actor-Critic}
Soft actor-critic (SAC) is a state-of-the-art off-policy deep reinforcement learning algorithm that has shown promise across a wide variety of continuous control tasks~\cite{sac1,sac2} and recently discrete action settings~\cite{sacd}. Unlike traditional reinforcement learning, SAC seeks to derive an optimal policy based on a maximum entropy objective function
\begin{equation}
    \pi^{*} = \argmax_{\pi}E[\sum_{t\geq 0}\gamma^{t}(r(s_{t}, a_{t}) + \alpha \mathcal{H}(\pi(\cdot|s_{t})))],
\end{equation}
where $\mathcal{H}(\pi(\cdot|s_{t}))$ represents the entropy of the policy distribution for a given state $s_{t}$ such that
\begin{equation}
    \mathcal{H}(\pi(\cdot|s_{t})) = E[-\log(\pi(\cdot|s_{t}))].
\end{equation}
The temperature parameter $\alpha$ represents the trade-off coefficient between expected returns and the entropy term. The standard RL objective is recovered when $\alpha \rightarrow~0$.

While SAC and SACD perform well in many environments, the performance is greatly sensitive to the choice of $\alpha$ and subsequently the target entropy. Recent work by \cite{te_anneal} introduced a target entropy annealing approach to address the sensitivity of the target entropy parameter. In this work, we adopt the use of target entropy annealing, which we found to be essential to obtain good performance in the air transportation environment.

\section{Approach}
\label{approach}
In this section, the aircraft separation assurance problem is introduced and formulated as an MDP by defining the state space, the action space, and the reward function. We then detail the distributed asynchronous training setup used to achieve a 10x increase in training throughput.

\subsection{Aircraft Separation Assurance}
Separation assurance involves preventing a loss of separation (LOS) event with aircraft in trail, at intersections, and at metering fixes by providing advisory maneuvers to aircraft. Any given aircraft in the environment is referred to as an ownship with all other aircraft in the airspace referred to as intruder aircraft from the ownship's point of view. In this way, each ownship will have its own associated intruder aircraft. The LOS threshold, $d^{\text{LOS}}$, defines a safety radius around ownship where operations with an intruder within the threshold become increasingly dangerous. Violating the loss of separation threshold may result in collisions between aircraft or near midair collisions (NMACs) that often result in drastic maneuvers from the aircraft. This task is traditionally performed by human air traffic controllers; however, novel automation techniques are required to safely scale to the expected magnitude of air traffic for AAM.

\label{deeprl}
\subsection{State Space}
In order to provide a scalable solution for increasing air traffic, we adopt a centralized training, decentralized execution scheme where each aircraft is considered an agent and with training, learns a cooperative policy for navigating through the airspace safely and efficiently. The state space for this environment consists of ownship information as well as information from the surrounding air traffic that is dynamic in size as aircraft take-off and land. 
The state is therefore decomposed into the ownship state and the intruder state. In order to handle variability in the intruder aircraft information, we then adopt the use of attention networks~\cite{luong2015effective}, similarly to the D2MAV-A algorithm~\cite{D2MAV_A}. This resulting intruder attention vector provides a fixed-length vector representation for network optimization. The ownship state space $s$ at time $t$ is defined as
\begin{multline*}
    s_{t} = 
    (\psi, z, \dot{v}_{x}, \dot{v}_{z} ,v_{x}, v_{z}, gs_{\text{East}}, gs_{\text{North}}, \\
    t, a_{t-1}, x_{\text{wpt}}(j)-x, y_{\text{wpt}}(j)-y) \quad \forall \; j \in [1, N_{\text{wpt}}],
\end{multline*}
to include heading ($\psi$), altitude ($z$), horizontal acceleration ($\dot{v}_{x}$), vertical acceleration ($\dot{v}_{z}$), horizontal speed ($v_{x}$), vertical speed ($v_{z}$), east and north ground speed ($gs_{\text{East}}$, $gs_{\text{North}}$), time of day ($t$), previous action ($a_{t-1})$ , and $N_{\text{wpt}}$ future ownship relative waypoint positions ($\bar{x}_{\text{wpt}}-x$, $\bar{y}_{\text{wpt}}-y$). $N_{\text{wpt}}$ is a hyperparameter that specifies how many future route segments to consider. The state space $h$ for the $i$ intruder aircraft available to the ownship at time $t$ is defined as 
\begin{multline*}
    h_{t}(i) = 
    (\psi^{(i)}_{\text{rel}}, z^{(i)}_{\text{rel}},\dot{v}^{(i)}_{x}, \dot{v}^{(i)}_{z}, v_{x}^{(i)}, v_{z}^{(i)}, gs^{(i)}_{\text{East}}, gs^{(i)}_{\text{North}}, \\ \phi^{(i)}, d^{(i)}_{o},  a^{(i)}_{t-1}, 
    x^{(i)}_{\text{wpt}}(j)-x, y^{(i)}_{\text{wpt}}(j)-y) \; \forall \; j \in [1, N_{\text{wpt}}]. 
\end{multline*}
The intruder state space contains information on the intruder aircraft similar to the ownship state with the intruder's acceleration, velocities, and previous action. It also includes ownship relative values of relative heading ($\psi^{(i)}_{rel}$), relative altitude ($z^{(i)}_{\text{rel}}$), relative bearing ($\phi^{(i)}$), the straightline distance between ownship and intruder ($d^{(i)}_{o}$), and the relative waypoint positions. 
With the state space specified, the components of the attention network can be defined as
\begin{equation}
    \label{eq:attention1}
    \text{score}(s_{t},\bar{h}_{t}) = 
    s^{\top}W_{1} \bar{h}_{t}\\
\end{equation}
\begin{equation}
    \label{eq:attention2}
    \eta_{s_{t},\bar{h}_{t}} = \frac{\text{exp}(\text{score}(s,\bar{h_{t}}))}{\sum_{j=1}^{n}\text{exp}(\text{score}(s_{t},\bar{h}^{j}_{t}))}
\end{equation}
\begin{equation}
    \label{eq:attention3}
    c_{s} = \sum_{i=1}^n\eta_{s,\bar{h}_{t}}\bar{h}^i_{t}
\end{equation}
\begin{equation}
    \label{eq:attention4}
    k_{s_{t}} = f(c_{s_{t}}) = \text{tanh}(W_{2}c_{s_{t}}),
\end{equation}
where Luong's multiplicative style \citep{luong2015effective} is used as the score calculation. $\eta_{s_{t},\bar{h_{t}}}$ is the attention weights of the ownship with respect to all of the other intruder aircraft, $c_{s}$ is the context vector that represents the weighted contribution of the surrounding air traffic, and $k_{s_{t}}$ is the attention vector that represents the abstract understanding of the surrounding air traffic. $W_{1}$ and $W_{2}$ are both learnable weight matrices determined through the neural network training. We then concatenate $k_{s_{t}}$ with $s_{t}$ to obtain the fixed length vector that can be passed through standard feed-forward layers of the actor and critic networks. Given that the attention network is operating as a state pre-processor, it can be used in both the actor and critic networks for SACD, or as part of a shared-layer network architecture.

\subsection{Action Space}
Actions for the agent reflect speed and altitude maneuvers, with a decision step of 4 seconds. The decision step is treated as a hyperparameter that can be modified based on the application. The action space is defined as
\begin{equation*}
    a_{t} = \{\dot{v}_{x_{-}}, \dot{v}_{x_{0}}, \dot{v}_{x_{+}}, v_{z_{-}}, v_{z_{0}}, v_{z_{+}} \}.
\end{equation*}
The available action are decrease speed ($\dot{v}_{x_{-}}$), maintain current speed ($\dot{v}_{x_{0}}$), increase speed ($\dot{v}_{x_{+}}$), descend ($v_{z_{-}}$), maintain altitude ($v_{z_{0}}$), and climb ($v_{z_{+}}$). Given that there are multiple vertically stacked air corridors, climb and descend actions are automatically stopped when reaching a new lane and the agent must re-select a climb or descend action to continue. The magnitude of the actions are dependent on the performance envelope for a given aircraft type. In addition, selected actions that result in speeds or altitudes outside of the performance envelope have no effect.

\subsection{Reward Function}
In the context of separation assurance, the primary objective is to maintain a safe distance from intruder aircraft; however, operational suitability objectives (e.g., minimize maneuvers) are also important for real-world deployment. We achieve this objectives through defining the reward function as
\begin{equation}
\label{reward_func}
R(s_t,h_t,a_{t}) = R(s_{t}, h_{t}) + R(a_{t}) - \Omega,
\end{equation}
where $R(s_{t}, h_{t})$ and $R(a_{t})$ are defined as
\begin{equation}
R(s_{t}, h_{t}) = 
    \begin{cases}
      \multirow{2}{*}{-1,} & \text{if $d^{c}_{o} < d_{x}^{\text{NMAC}}$}\\
                           & \text{and } z_{rel} < d_{z}^{\text{NMAC}}\\
      -\chi + \delta \cdot d^{c}_{o}, & \text{if $d^{\text{NMAC}} \leq d^{c}_{o} < d^{\text{MAX}}$}\\
      0, & \text{otherwise}
    \end{cases},  
\end{equation}
\begin{equation}
R(a_{t}) = 
    \begin{cases}
      0, & \text{if $a_{t} \in $ [$\dot{v}_{x_{0}}, v_{z_{0}}$]}\\
      -\epsilon, & \text{if $a_{t} \in $ [$\dot{v}_{x_{-}}, \dot{v}_{x_{+}}$]}\\
      -\lambda, & \text{if $a_{t} \in $ [$ v_{z_{-}}, v_{z_{+}}$]}\\
    \end{cases}.
\end{equation}

In $R(s_{t}, h_{t})$, $d^{c}_{o}$ is the distance from the ownship to the closest intruder aircraft and $d^{\text{MAX}}$ is the maximum distance to consider the closest intruder aircraft in the reward function. The hyperparameters $\chi$ and $\delta$ are small, positive constants to penalize aircraft as they approach the separation threshold, $d^{\text{NMAC}}$. In $R(a_{t})$, $\epsilon$ and $\lambda$ represent penalties for advisories that require a deviation from the aircraft's current speed or altitude, respectively, to encourage the agent to minimize maneuvering  actions. Finally, in $R(s_t,h_t,a_{t})$, the hyperparameter $\Omega$ represents a small, positive constant that is applied at every step in scenario. This parameter discourages aircraft from airborne holding, since slower aircraft will incur the $\Omega$ penalty for an extended time. Table~\ref{use_case_param} displays the finalized use-case hyperparameters.

\begin{table}[bt]
\caption{Finalized use-case hyperparameters.}
\centering
\label{use_case_param}
\begin{tabular}{lll}
Parameter                    & Value &  \\ \hline
\multicolumn{1}{l|}{$d_{x}^{\text{NMAC}}$}  & 500 feet   \\
\multicolumn{1}{l|}{$d_{z}^{\text{NMAC}}$}  & 100 feet   \\
\multicolumn{1}{l|}{$v_{x}: \text{range}$} &  [5, 65] knots    &  \\
\multicolumn{1}{l|}{$z: \text{range}$} &  [400, 1600] feet    &  \\
\multicolumn{1}{l|}{$N_{\text{wpt}}$} &  5    &  \\
\multicolumn{1}{l|}{$d^{\text{MAX}}_{o}$} &  3280 feet     &  \\
\multicolumn{1}{l|}{$\gamma$} &  0.99     &  \\
\multicolumn{1}{l|}{Reward coefficient $\chi$} &  0.1     &  \\
\multicolumn{1}{l|}{Reward coefficient $\delta$} &  0.0001     &  \\
\multicolumn{1}{l|}{Reward coefficient $\epsilon$} &  0.001     &  \\
\multicolumn{1}{l|}{Reward coefficient $\lambda$} &  0.01     &  \\
\multicolumn{1}{l|}{Reward coefficient $\Omega$} &  0.001     &  \\
\multicolumn{1}{l|}{Replay memory capacity} &  8000000     &  \\
\multicolumn{1}{l|}{Learning rate} &  5e-5     &  \\
\multicolumn{1}{l|}{Batch size} &  512     &  \\
\hline
\end{tabular}
\end{table}

\subsection{Distributed Asynchronous Training}
\begin{figure}[t]
\begin{center}
\centerline{\includegraphics[width=0.9\columnwidth]{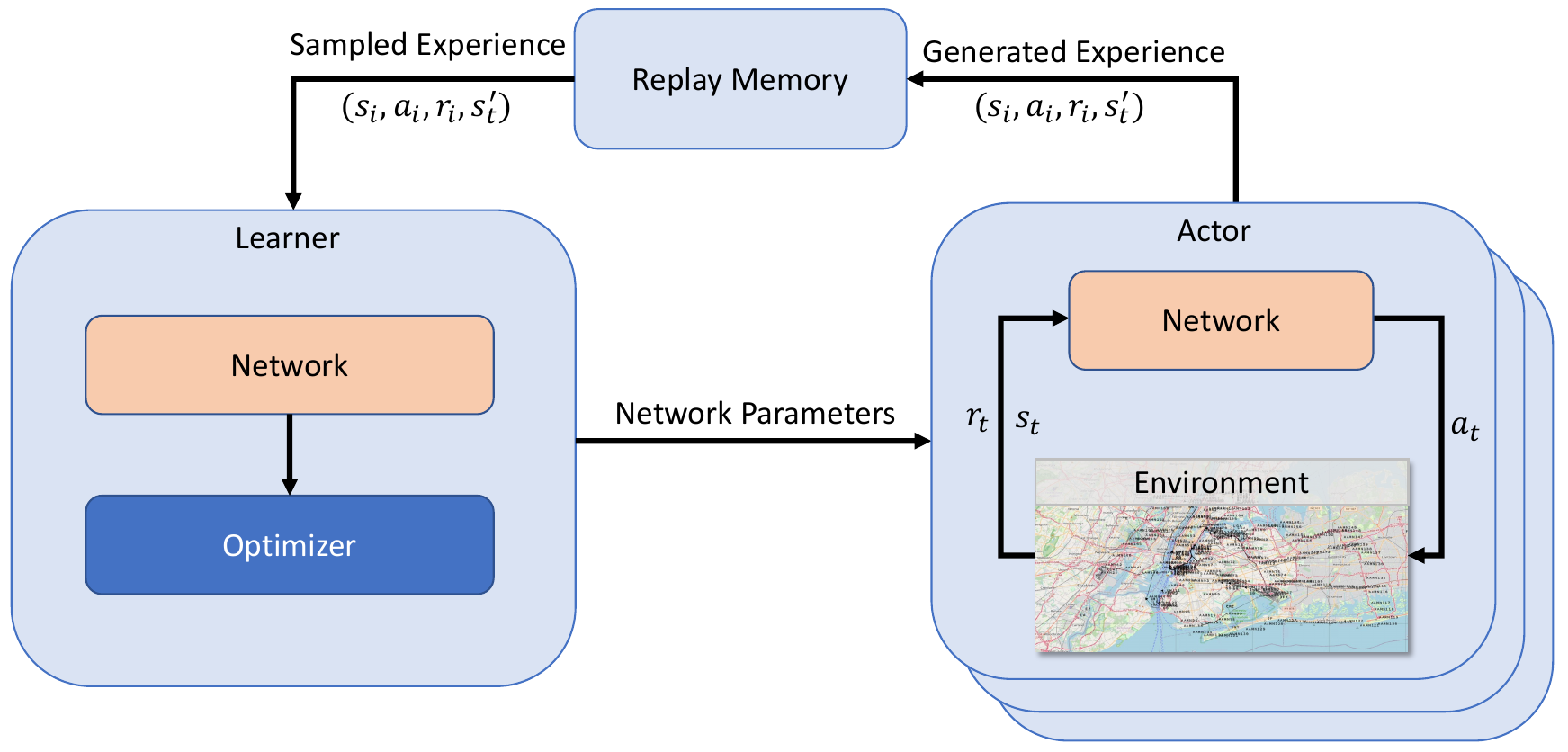}}
\caption{Distributed asynchronous training architecture.}
\label{training_setup}
\end{center}
\end{figure}

D2MAV-A introduced a distributed synchronous training procedure where parallel actors collect state-transition experience from the environment for a centralized learner to train on. Given that D2MAV-A is an on-policy RL algorithm based on Proximal Policy Optimization~\cite{schulman2017proximal}, a synchronization step is required to ensure that each actor is collecting experience from the most up-to-date policy.
SACD, an off-policy RL algorithm, does not require each actor to be collecting experience with the latest policy. Therefore, we can then decouple the algorithm training from the algorithm execution in the environment. In this way, the algorithm training can be performed asynchronously from the distributed actors to achieve high training throughput. Figure~\ref{training_setup} illustrates the asynchronous centralized learning, decentralized execution scheme. We adopt the same network architecture for SACD as in \cite{sacd} with the addition of the attention network for state preprocessing. The network layers consisted of 256 nodes for both the actor and critic networks. For target entropy annealing, we use the parameter valeus introduced in \cite{te_anneal}, with the exception of the standard deviation threshold, which we set to 0.07. Experiments were performed on the Lincoln Laboratory Supercomputer, consisting of 16 Intel Xeon Gold 6248 2.5 Ghz compute nodes with two NVIDIA Tesla V100 graphics processing units (GPUs) per compute node~\cite{reuther2018interactive}. 

\section{Use-Case: Urban Air Corridors}
\label{problem}
Near-term AAM operations are expected to leverage existing visual flight rule (VFR) route networks (e.g., helicopter routes) as they will largely represent AAM corridors at lower air traffic densities~\cite{faa_conops}. As such, we developed an environment based on the VFR route network and 29 vertiport locations for New York City as presented in \cite{doi:10.2514/6.2021-2381} with the addition of vertically stacked lanes at 400, 700, 1000, 1300, and 1600 feet. The following subsections discuss the scenarios designed for this use-case, baselines, and experiment setup.

\subsection{Scenario Design}
A total of 20 days of representative AAM traffic is generated by UAMToolkit \cite{doi:10.2514/6.2021-2381} with varying fleet sizes to provide a diverse set of operational densities. The AAM traffic demand is based on a displacement of 5\% of the New York City taxi cab market, subject to the number of available AAM aircraft.
The scenario generation takes into account imperfect aircraft altitude by adding noise in the form of a uniform distribution with a minimum value of -100 ft and maximum value of 100 ft to offset the selected initial altitude. The fleet size determines how many aircraft are available to operate simultaneously; however, simultaneous operations will be limited based on the availability of vertiport parking spots. In this study, each vertiport was assumed to have four parking spots which results in a max of approximately 200 simultaneous operations. If the fleet size exceeds 200 aircraft, overall network utilization increases due to lower idle times by aircraft repositioning for passenger pickup. For this use-case, two aircraft performance models were chosen to simulate flights: (1) Eurocopter EC-135 and (2) surrogate AAM vehicle based on publicly available AAM aircraft specifications.

\subsection{Baselines}
Two baselines were used in this study to benchmark the proposed SACD-A algorithm: (1) an unequipped agent that does not implement any separation commands (i.e., aircraft follows original flight plan) and (2) the D2MAV-A algorithm with speed and vertical lane change commands, an extension of the original speed-only D2MAV-A implementation.

\subsection{Experiment Setup}
Using the AAM-Gym testbed~\cite{9925762} with 40 parallel workers for policy-rollout, each algorithm was trained for 10,000 iterations where one iteration is 64 environment steps (4 simulation seconds).
Following the completion of training, evaluation was performed on 100 randomly sampled 3-hour windows of AAM operations for a given day. A sensitivity analysis was performed to understand how robust the algorithms are to state observation noise and fleet size by simulating various fleet sizes with both automatic dependent surveillance-broadcast (ADS-B) noise and perfect surveillance. ADS-B noise is applied as a Gaussian distribution over the latitude, longitude, and altitude, with a standard deviation error magnitude of 0.0001 for latitude and longitude, and 100 feet for altitude. To test the robustness of the algorithms under various levels of uncertainty, we performed a sweep over two stressing parameters : (1) probability of communication and (2) probability of policy equipage with a fleet size of 100 aircraft. Probability of communication refers to the ability of a given aircraft to receive intruder state information. This condition is applied at aircraft initialization such that an aircraft without communication does not observe the intruder aircraft for the entire flight duration. Probability of policy equipage represents the likelihood that an aircraft is equipped with a separation assurance logic to minimize the loss of separation with the surrounding aircraft. Policy equipage is determined at aircraft initialization and aircraft not equipped with the logic follow their original flight altitude and speeds (identical to unequipped aircraft). In situations containing unequipped aircraft, the algorithms must learn and adapt to the non-cooperative aircraft.

\begin{figure}[t]
\begin{center}
\centerline{\includegraphics[width=0.9\columnwidth]{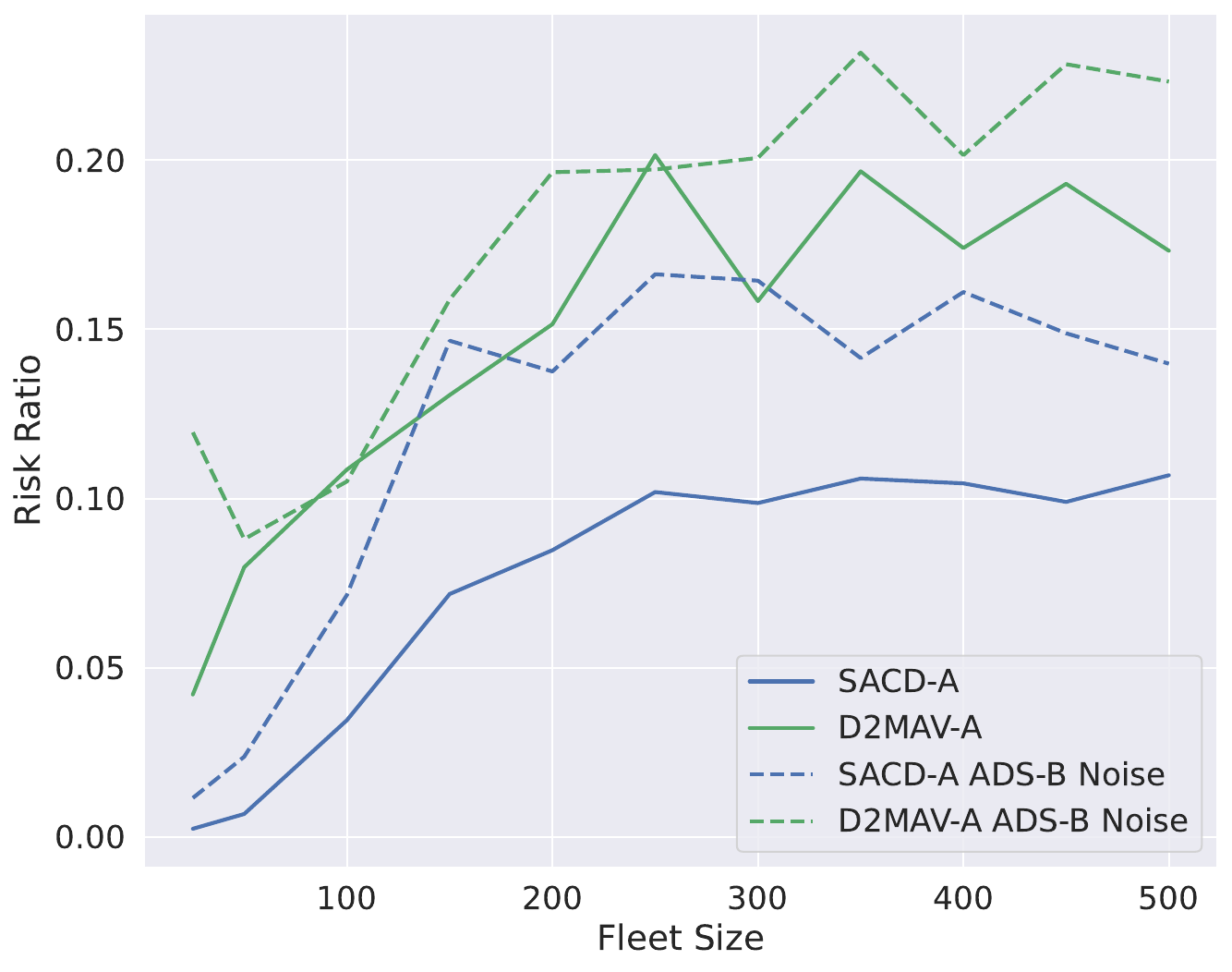}}
\caption{Risk ratio versus fleet size.}
\label{fleet}
\end{center}
\end{figure}

\section{Results}
\label{results}
We adopt the use of the risk ratio as the primary metric for evaluating the safety of the algorithms. Risk ratio is a commonly used metric for aircraft collision avoidance systems~\cite{9081631} and is defined as
\begin{equation}
\text{risk ratio} = \frac{P(\text{NMAC})_{\text{logic}}}{P(\text{NMAC})_{\text{no logic}}}.
\end{equation}
The risk ratio provides a measure of how much safety improvement can be achieved using an algorithm compared to the case when no logic is present (unequipped aircraft). Values close to zero represent that the algorithms resolved all NMAC events, whereas values greater than one indicate the algorithms results in more NMACs than the unequipped aircraft. Risk ratio equal to one indicates no safety improvement with the algorithms.

Figure~\ref{fleet} shows the algorithm risk ratio for various fleet sizes and surveillance sources. Both algorithms were able to reduce airspace risk over the unequipped agent given risk ratio values less than one. For all fleet sizes SACD-A outperforms D2MAV-A, achieving a steady-state risk ratio at fleet~size = 250. Interestingly, SACD-A with ADS-B noise outperforms D2MAV-A with perfect surveillance, demonstrating the robustness of SACD-A.

\begin{figure}[t]
\begin{center}
\centerline{\includegraphics[width=0.9\columnwidth]{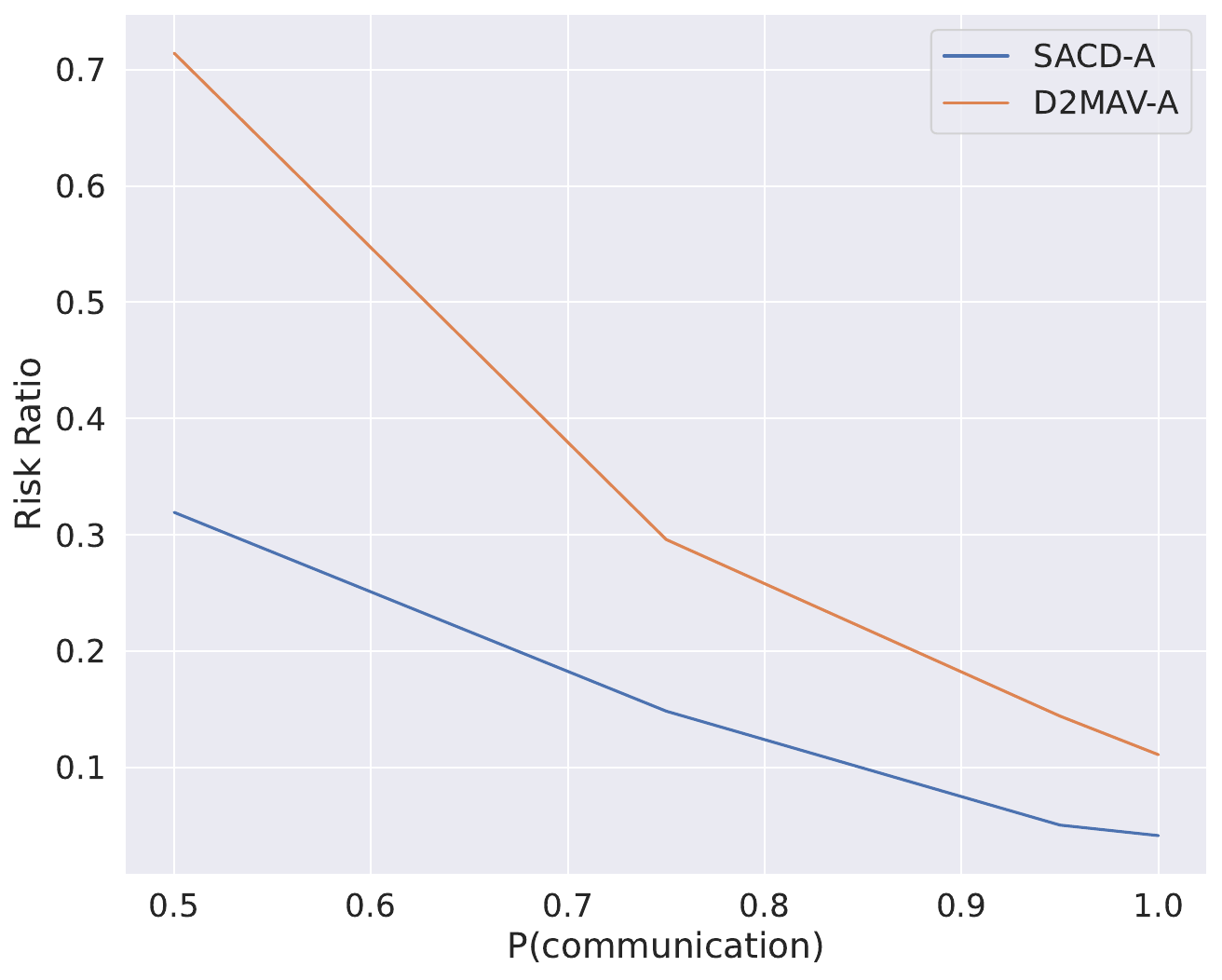}}
\caption{Risk ratio versus probability of communication.}
\label{pcomms}
\end{center}
\end{figure}

Figure~\ref{pcomms} shows the algorithm risk ratio for various communication probabilities. As expected, risk ratio increases as the probability of communication decreases since fewer agents are able to receive intruder state information. However, it is seen that SACD-A achieves a much smaller risk ratio over D2MAV-A, with the difference becoming more significant as the probability of communication decreases.

\begin{figure}[t]
\begin{center}
\centerline{\includegraphics[width=0.9\columnwidth]{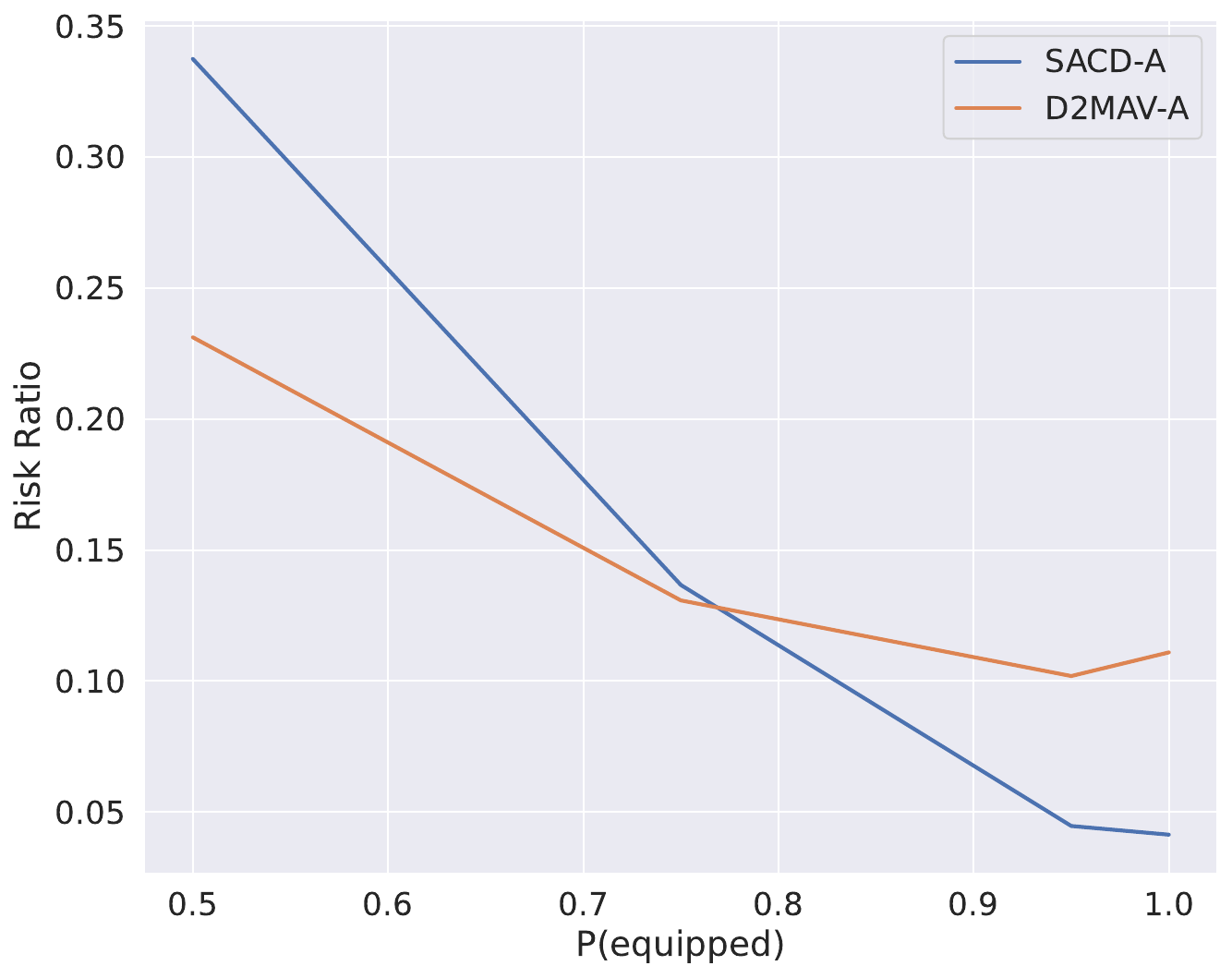}}
\caption{Risk ratio versus probability of equipped.}
\label{pequip}
\end{center}
\end{figure}

The impact of the probability of equipage is shown in Figure~\ref{pequip}. At $P(\text{equipped})\leq0.75$, D2MAV-A outperforms SACD-A, indicating that D2MAV-A may perform better at adapting to the behavior of unequipped agents. However, when a greater majority of the aircraft are logic-equipped ($P(\text{equipped})>0.75$) SACD-A significantly outperforms D2MAV-A, resulting in a risk ratio of 0.05 when at least 95\% of aircraft are equipped with the SACD-A policy.

Thus far the results have focused on algorithm safety, however it is also important to consider the operational suitability of the learned policy (e.g., reduce unnecessary maneuvers). Table~\ref{dist} shows the probability distribution of aircraft maneuvers for the SACD-A algorithm from the evaluation scenarios. It is seen that the majority of actions selected by the agent are non-maneuvering actions including `maintain speed' ($\dot{v}_{x_{0}}$) and `maintain altitude' ($v_{z_{0}}$), successfully achieving our objective of minimizing maneuvering to only when it is necessary. The impact of the step penalty is reflected in the percentage of `speed up' ($\dot{v}_{x_{+}}$) actions, given the agent is encouraged to navigate the route as quickly as possible. From the results, it is shown that the `climb' action was preferred over `descend', however there was no preference in the reward function for choosing between climb or descend. We attribute this behavior to the initial altitude distribution being slightly skewed to lower altitudes.
\begin{table}[bt]
\centering
\begin{tabular}{ccccccc}
     \multicolumn{1}{c}{$a$}            & $\dot{v}_{x_{(-)}}$ & $\dot{v}_{x_{(0)}}$ & $\dot{v}_{x_{(+)}}$ & $v_{z_{(-)}}$ & $v_{z_{(0)}}$ & $v_{z_{(+)}}$ \\ \hline
\multicolumn{1}{c|}{$P(a)$} & 0.005 &  0.422   &  0.05  & 0.028  & 0.488  & 0.007 
\end{tabular}
\caption{SACD-A action distribution throughout evaluation.}
\label{dist}
\end{table}

The algorithm sample efficiency is compared over the 10,000 training iterations. SACD-A trained on 2.36 billion transitions, while D2MAV-A trained on 256 million transitions, an almost 10x increase. This shows that the distributed asynchronous training architecture for SACD-A provides increased training throughput compared to D2MAV-A.

\section{Discussion}
\label{discussion}
The results of this work provide a promising solution to increase safety and efficiency of air transportation not only in AAM, but also in today's commercial aviation. When considering a safety-critical application such as separation assurance, there are additional steps that would need to be performed to transition this approach to real-world deployment. First, the simulation environment would need to introduce more accurate models to depict the operating environment, such as action latency and weather. After obtaining a suitable environment of the real world, a standards community would need to be formed in order to validate the safety and efficacy of the proposed algorithm. Existing systems such as the ACAS X collision avoidance system (a dynamic programming approach) approved by RTCA and the FAA provide a model for achieving community approval, however this approach will need to be modified for neural-network based algorithms. In general, having an in-depth understanding of the learning-based system's behavior and leveraging validation/verification techniques such as adaptive stress testing to identify failure modes of learning-based systems will be required before full deployment in the real world.

In the near-term, for today's commercial aviation sector, deployment could be achieved through human supervisory control. In this setting, the algorithm instead acts as a recommender system to provide air traffic control a recommended maneuver to resolve a potential conflict or improve efficiency. However, further simulation studies and human-in-the-loop experiments would need to be conducted to understand the impact of real-world implementation.

\section{Conclusions}
\label{conclusion}
A decentralized reinforcement learning framework is introduced to safely and efficiently separate aircraft in high density AAM corridors through speed and vertical maneuvers. A rigorous set of numerical experiments demonstrate the effectiveness of the SACD-A framework over existing approaches by introducing sources of uncertainty and high traffic density scenarios. The operational suitability of the proposed framework is shown through maximizing non-maneuvering actions, so that actions are selected only if necessary to resolve a conflict or increase efficiency. Looking forward, we to plan apply the SACD-A framework to more complex scenarios as well as to explore the framework interoperability with existing airspace deconfliction systems.

\fontsize{9.0pt}{10.0pt} \selectfont

\end{document}